\documentclass{article}
\usepackage[preprint]{neurips_2026}

\usepackage[utf8]{inputenc}
\usepackage[T1]{fontenc}
\usepackage{hyperref}
\usepackage{url}
\usepackage{booktabs}
\usepackage{amsfonts}
\usepackage{amsmath}
\usepackage{nicefrac}
\usepackage{microtype}
\usepackage{xcolor}
\usepackage{graphicx}
\usepackage{caption}
\usepackage{subcaption}
\usepackage{float}
\usepackage{afterpage}
\usepackage{enumitem}
\graphicspath{{figs/}}

\hypersetup{pdfauthor={Louis Yiven Zhu},
  pdftitle={One Capability or Many? Structural and Predictive Tests of Benchmark Validity Disagree About Economic Benchmarks for Frontier AI},
  pdfsubject={Construct validity of economic benchmarks on frontier AI leaderboards},
  pdfkeywords={construct validity; benchmark validity; incremental validity; factor analysis; leaderboard audit; large language models},
  colorlinks=true,linkcolor=blue,citecolor=blue,urlcolor=blue}

\newcommand{\Rsq}{R^{2}}
\newcommand{\dmse}{\Delta\mathrm{MSE}}
\newcommand{\tauthree}{$\tau^{3}$-Banking}
\newcommand{\tautwo}{$\tau^{2}$-Bench}

\title{One Capability or Many?\\Structural and Predictive Tests of Benchmark Validity\\Disagree About Economic Benchmarks for Frontier AI}

\author{%
  Louis Yiven Zhu\thanks{ORCID 0009-0001-5579-0340.} \\
  University of Oxford \\
  \texttt{yiven.zhu@oii.ox.ac.uk} \\
}

\begin{document}

\maketitle

\begin{abstract}
Frontier-model leaderboards now rank systems on economic benchmarks, and those rankings inform what organisations buy and what regulators scrutinise. Whether such benchmarks measure a capability distinct from general test-taking is a question of construct validity that a structural test and a predictive test can answer in opposite ways. We show that they do on a hash-pinned snapshot of a frontier leaderboard with 421 model configurations across twelve benchmarks, four of them economic, of which 103 configurations carry all three sparsely scored economic benchmarks and 96 carry all twelve; four hypotheses and their thresholds were fixed before analysis, and every deviation from the plan is reported. The first factor of a three-factor extraction carries 74.5\% of common variance and tracks release date ($\Rsq = 0.505$), and date adjustment lowers its share by 14.9 points. Under the dimensionality rule fixed in advance the economic benchmarks form no factor of their own. A leave-one-benchmark-out test with factors re-estimated inside every fold nevertheless finds that a multi-factor representation predicts held-out economic scores better than a single general index (pooled $\dmse$ 0.037, 95\% bootstrap interval $[0.019, 0.055]$) under the linear learners that fit best, an advantage that reverses for tree learners. Under the linear learners the same representation also predicts the eight other benchmarks better, so the battery carries predictive structure that one index misses and the economic benchmarks share it without forming a distinct factor. Construct validity should therefore be assessed by predictive tests alongside structural ones. We give a two-test protocol for benchmark builders and release the pinned data, the analysis plan and the code.
\end{abstract}

\section{Introduction}
\label{sec:intro}

Public leaderboards have become the dominant instrument for comparing and procuring frontier language models, and alongside academic knowledge tests they now carry \emph{economic} benchmarks, suites built to proxy paid professional work and multi-step tool use. Whether those benchmarks measure a capability distinct from general test-taking is a question of construct validity, the degree to which a measurement captures the construct it claims to \citep{cronbach1955,messick1998}, and the two standard tests of construct validity, one structural and one predictive, can disagree about the same benchmarks. The structural test asks whether a group of benchmarks forms a factor of its own once the variance they share with every other benchmark is accounted for, the discriminant criterion of \citet{campbell1959}. The predictive test asks whether the rest of the battery predicts them better out of sample through a richer representation than through a single general index, an adaptation of the incremental-validity criterion of \citet{sechrest1963} to a battery without an external criterion. On a snapshot of one frontier leaderboard the structural test says no and the predictive test says yes. Under a dimensionality rule fixed before analysis the economic benchmarks form no distinct factor, yet under the linear learners that fit best a multi-factor representation predicts held-out economic scores better than a general index, by a modest margin with an interval that excludes zero. The advantage extends to the other eight benchmarks, so the disagreement concerns how many dimensions the battery carries. This paper is about that disagreement, what produces it, and what it implies for how the validity of economic benchmarks should be judged.

The question has stakes because of who reads leaderboards and what they now list. Alongside GPQA (Graduate-Level Google-Proof Q\&A) \citep{rein2023gpqa} and Humanity's Last Exam (HLE) \citep{phan2025hle}, the leaderboard studied here carries GDPval \citep{patwardhan2025gdpval}, the $\tau$-Bench family \citep{yao2024taubench,barres2025tau2}, and terminal or agentic coding suites \citep{merrill2026terminalbench}, and rankings on these columns inform what organisations buy and what regulators scrutinise. A separate column is warranted only if they measure something a single ``intelligence'' score does not. Machine learning has been urged to hold its benchmarks to that standard \citep{raji2021,jacobs2021,bowman2021}, audits find construct validity under-examined in AI evaluation \citep{bean2025,kearns2026}. What remains open is whether an economic benchmark adds anything a general index lacks when the held-out column never enters estimation, and whether the structural and predictive answers to that question agree. The question is open because the nearest study compares individual benchmarks pairwise over ten models (Section~\ref{sec:related}), and the data and methods to answer it at leaderboard scale exist.

\begin{figure}[!t]
\centering
\includegraphics[width=\textwidth]{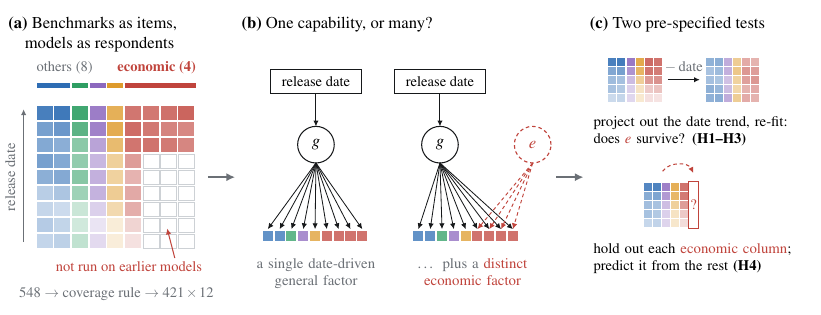}
\caption{\textbf{Study design.} (a) Benchmarks as items (columns, economic in red) and model configurations as respondents (rows, ordered by release date); a coverage rule from the analysis plan reduces 548 configurations to a $421 \times 12$ matrix. (b) The two candidate structures, $g$ alone or $g$ with a distinct economic factor $e$. (c) The two pre-specified tests, date adjustment then re-fitting (H1--H3) and leave-one-benchmark-out prediction (H4); grids are stylised.}
\label{fig:pipeline}
\end{figure}

We frame a snapshot of frontier models scored across twelve benchmarks as a psychometric problem, with benchmarks as items, models as respondents, and scores as the responses from which latent structure is inferred, and answer it under a pre-specified design. Figure~\ref{fig:pipeline} shows the score matrix (a), the two candidate structures (b) and the two tests (c). Four research questions organise the analysis, each mapping to one pre-specified hypothesis (Section~\ref{sec:hyp}); the first three concern structure (Task~1, unsupervised) and the fourth prediction (Task~2, supervised).
\begin{enumerate}[label=\textbf{RQ\arabic*}, leftmargin=3.2em, itemsep=0pt, topsep=2pt, parsep=0pt]
\item \textbf{(dominance)} How many capabilities does the battery measure, and how dominant is the first?
\item \textbf{(release date)} How much of the leading axis is a release-date trend, and how much survives date adjustment?
\item \textbf{(structural distinctiveness)} After date adjustment, do the economic benchmarks load on a factor of their own?
\item \textbf{(predictive validity)} Does a multi-factor representation beat a single general index on held-out economic scores?
\end{enumerate}

The paper makes four contributions. First, it shows that a structural and a predictive test of construct validity return opposite verdicts on the same economic benchmarks, and it argues from that dissociation that validity claims for a benchmark should say which test they rest on. Second, it introduces a leakage-safe leave-one-benchmark-out predictive test against a general-index baseline and finds that, under linear learners, the multi-factor advantage holds across the battery. Third, it measures how much of the dominant axis tracks model release date, which on a frontier snapshot carries what scale carries across generations. Fourth, it delivers the analysis under pre-specified hypotheses and thresholds with hash-pinned inputs, documenting every deviation and reporting the two failed tests as failed. Section~\ref{sec:related} places the study in the literature, Sections~\ref{sec:data} and~\ref{sec:methods} give the data and the two tasks, Section~\ref{sec:results} the results by hypothesis, and Section~\ref{sec:discussion} what the disagreement implies.

\section{Related work}
\label{sec:related}

Two strands of the construct-validity programme bear on this paper. The first treats LLM scores as psychometric data, following the case for measuring artificial systems with the instruments built for measuring minds \citep{hernandez2017} and the century-old observation that cognitive tests yield a positive manifold and a general factor \citep{spearman1904}. \citet{ilic2024} recover that manifold across 591 models on twelve tests, with a mean inter-test correlation of 0.73 and a general factor correlating with parameter count at about 0.6. How many further factors accompany the general one depends on the retention rule as much as on the data. \citet{burnell2023}, on 29 LLMs and 27 tasks, resolve capability into three factors; \citet{maimon2026} recover eight from 60 models on 44 benchmarks under Kaiser's rule and a cumulative-variance threshold. That sensitivity is one reason to pair a structural test with a predictive one. \citet{ruan2024} use principal components of a score matrix as a capability space in which performance, agentic performance included, becomes predictable \citep[see also][]{zhou2025scales}. \citet{ho2025rosetta} stitch 38 benchmarks onto a single capability scale over 179 models, which is the one-dimensional assumption our single-index baseline embodies. Closest to our design, \citet{kearns2026} makes the validity question quantitative on the Open LLM Leaderboard, where a plain factor analysis fits a single factor explaining 72\% of communal variance that tracks parameter count. Residualising every score on a fitted scaling law then lowers the mean inter-item correlation from 0.64 to 0.48, and that residualisation is the template for our date adjustment. Our design departs from his in three respects. The population is a frontier snapshot on which parameter counts are mostly unpublished, so release date takes the role scale plays there; the economic benchmarks are isolated as a block with a pre-specified discriminant test; and a leave-one-benchmark-out predictive test sits beside the structural one, which his study does not run. Across all of these studies scale is the variable to control; on a frontier snapshot we find that release date takes that role and compute adds little once date is removed, which suggests the variable to adjust for has moved with the field.

The second strand audits benchmarks as artefacts. Calls to repair benchmarking predate the frontier era \citep{bowman2021,kiela2021}, and holistic evaluation frameworks \citep{liang2022helm}, reproducibility guidance \citep{biderman2024}, and proposals for a full evaluation science \citep{weidinger2025} have since sharpened them. Reviewing 445 benchmarks, \citet{bean2025} find that 53.4\% present any evidence for the construct validity of their benchmark and 16.0\% report uncertainty estimates or statistical tests when comparing models. On the temporal side, \citet{akhtar2026} find that nearly half of sixty benchmarks are saturated and that saturation rises with benchmark age, which makes release date a first-order variable whenever capabilities are compared across a moving frontier. Closest in question, \citet{wan2026econevals} introduce occupation-level benchmarks and test whether they predict GDPval scores better than generic capability benchmarks do, comparing Spearman correlations benchmark by benchmark across ten models, and find that their benchmarks win or tie in most comparisons. Our test differs in what it holds fixed. The baseline is a general index built from the whole battery where EconEvals compares one benchmark against another; the factors are re-estimated inside every fold so the held-out column never enters estimation; the thresholds were fixed before analysis; and the sample is a hash-pinned snapshot of 421 configurations. None of this work pairs a structural test with an out-of-sample predictive one on the same data, and none asks whether the two agree.

\section{Data and design}
\label{sec:data}

Every number below rests on one hash-pinned snapshot and one coverage rule fixed before any score was inspected. The primary data set is a snapshot of the Artificial Analysis model leaderboard \citep{aa2026}, captured on 6 July 2026 directly from the public leaderboard page and pinned by its SHA-256 hash (Appendix~\ref{app:provenance}). The leaderboard is run by one independent operator that evaluates every model under a single harness on a fixed battery and publishes per-benchmark scores beside an aggregate index, so scores are comparable across providers. The snapshot contains 548 model configurations with per-benchmark scores, release dates, and provider metadata, where a configuration is one base model under one reasoning or effort setting. A secondary source, the Epoch AI Notable AI Models data set \citep{epoch2026}, supplies training-compute figures for a subset of models and serves only a scale robustness check. Both snapshots are pinned, so every number here is reproducible from fixed inputs.

To avoid selecting benchmarks or models on the very outcomes under study, we fixed the coverage rule in advance. A benchmark is retained when at least sixty models carry a score, and a model when it carries at least eight of the thirteen sufficiently covered benchmarks. Both floors were set as indicative values in the analysis plan and confirmed unchanged at the coverage audit; that confirmation used coverage counts alone, and no score, correlation or factor solution entered it. Appendix~\ref{app:coverage} shows what each protects. Under this rule thirteen of fourteen candidates pass; we drop APEX-Agents \citep{vidgen2026apex} for sparsity and demote MMMU-Pro \citep{yue2025mmmupro} to a sensitivity-only role, since it is the battery's only multimodal benchmark and requiring it would cut the complete-case grid from 96 to at most 66 models (Appendix~\ref{app:coverage}), which leaves twelve primary benchmarks and 421 model configurations. Four of the twelve are economic benchmarks proxying paid professional or agentic work, and eight span academic knowledge, scientific coding \citep{tian2024scicode,zhu2025critpt}, long-context retrieval, and instruction following \citep{pyatkin2025ifbench} (Appendix~\ref{app:battery}). The 421 rows are configurations, since one base model can appear under several reasoning or effort settings; the plan deduplicates to one row per base model as robustness check R1, and we keep the configuration level as primary because it is the more conservative choice.

Because the economic benchmarks are scored on far fewer models than the academic core, several nested samples appear in what follows, and each is named with its denominator. Of the 548 configurations parsed, 421 pass the coverage rule; 103 of those carry all three sparsely scored economic benchmarks (GDPval, Terminal-Bench v2.1, and \tauthree{}), the count on which the coverage audit gates; and 96 carry all twelve benchmarks. That complete-case grid of 96 is the only sample on which the economic block is complete, and it carries every hypothesis and the predictive test. The dense grid of 409 configurations scored on the nine near-universal benchmarks serves only the corroborating clustering (Appendix~\ref{app:clustering}), where sample size matters more than economic completeness. The 96-model grid is the set of frontier configurations on which the economic benchmarks have actually been run, and the claims below extend to that set and to no wider population.

Since the benchmarks are reported on incommensurable scales, mostly accuracy fractions alongside one hallucination-penalised score and GDPval's Elo rating (Appendix~\ref{app:battery}), we standardise each benchmark to zero mean and unit variance before any analysis. In the supervised task the standardisation is computed on each training fold and applied to its held-out fold, so no held-out information enters the fit. Missingness is structural, so we use complete-case analysis and impute nothing. On the complete-case grid the Kaiser--Meyer--Olkin (KMO) measure of sampling adequacy is 0.933, well above the 0.8 threshold \citep{kaiser1974}, and Bartlett's test of sphericity is decisively rejected \citep{bartlett1950}, so factor analysis is appropriate. The battery is also highly collinear before any modelling, with a mean off-diagonal Spearman correlation of $\rho = 0.79$ and every benchmark pair positive, and the economic benchmarks correlate with the academic core almost as strongly as with one another (Appendix~\ref{app:eda}). The manifold is therefore at least as strong here as the 0.73 that \citet{ilic2024} report on a broader population. The scope is internal validity, the correlational and predictive structure of aggregate scores on one cross-section; item-level responses, external economic impact, and saturation dynamics lie outside it.

\section{Methods}
\label{sec:methods}

\subsection{Task 1: latent structure}
\label{sec:task1}

Each structural question has its own quantity, and Appendix~\ref{app:derivations} states each method formally. The first factor's share of common variance measures dominance (RQ1), the change in that share after date adjustment measures the release-date component (RQ2), and the date-adjusted loading pattern tests distinctiveness (RQ3). To fix dimensionality we use Horn's parallel analysis \citep{horn1965}, which retains factors whose eigenvalue exceeds the 95th percentile of eigenvalues from random data of the same shape, with the scree criterion \citep{cattell1966} as a cross-check. On collinear data this rule is more conservative than the Kaiser rule \citep{kaiser1960} and than information criteria (Appendix~\ref{app:derivations}), and both rules retain one factor here. We prefer factor analysis to the principal-component description of \citet{ruan2024} because components mix common and unique variance, whereas the common-factor model isolates shared capability, which is the distinction a validity question turns on. The model is accordingly a maximum-likelihood exploratory factor analysis (EFA) with an oblique oblimin rotation \citep{jennrich1966,thurstone1947}, chosen because the date-adjusted factors are strongly correlated (Section~\ref{sec:h3}) and an orthogonal rotation would spread shared variance across artificially independent axes. Every factor is oriented so that higher means more capable. The estimators used throughout, this factor model, Thurstone factor scores, and the leave-one-benchmark-out prediction of Section~\ref{sec:task2}, are implemented in a released and versioned library, benchprobe v0.1.0 \citep{benchprobe2026}, whose acceptance tests lock the values reported below against the archived tables (Appendix~\ref{app:provenance}).

To separate a general capability from a shared time trend, we regress each standardised benchmark on release date, and on a subsample also on training compute, and re-fit the factor analysis on the residuals; we call this date adjustment. The adjustment does not treat time as noise. Later models arrive with newer data, algorithms and post-training, so the date-aligned component contains real progress, and the adjustment asks what the benchmarks measure beyond what the calendar already indexes. It removes from the benchmark covariance the rank-one component aligned with date (Appendix~\ref{app:derivations}), so the resulting change in the first factor's variance share measures how much of the apparent general factor the calendar already indexes, the analogue of the scaling-law residualisation \citet{kearns2026} reports. Clustering corroborates the structure without assuming a factor model (Appendix~\ref{app:clustering}).

\subsection{Task 2: predictive validity}
\label{sec:task2}

Task~2 recasts the distinctiveness question as a forecast, asking whether a model's score on a held-out economic benchmark follows from its behaviour on the others, a sharper test than the in-sample capability space of \citet{ruan2024} because the target column is withheld from estimation entirely. To that end, the leave-one-benchmark-out (LOBO) protocol takes each economic benchmark as the target in turn and predicts it from representations of the remaining eleven on the $n = 96$ complete-case grid, so that success demands transfer across benchmarks. Rung (i) of a five-rung predictor ladder uses release timing and scale only, rung (ii) a single mean-score general index, rung (iii) the first factor only, rung (iv) the $k$ factors (the three-factor solution of Section~\ref{sec:h3}), and rung (v) the $k$ factors plus model covariates. The contrast of interest sets rung (iv) against the single-index rung (ii).

To see whether the ladder comparison depends on model class, we fit each rung with four learners spanning the linear-regularised and tree families \citep{hastie2009}, ridge \citep{hoerl1970}, elastic net \citep{zou2005}, random forest \citep{breiman2001}, and gradient boosting \citep{friedman2001}. Hyperparameters are tuned by grid search in an inner five-fold cross-validation nested inside an outer five-fold split that supplies the reported out-of-fold error, the nesting that removes selection bias from the estimate \citep{varma2006} (grid in Appendix~\ref{app:hyper}). The plan does not fix the learner for the H4 contrast. We report it ridge against ridge, the learner with the lowest out-of-fold error on both rungs, and Appendix~\ref{app:robust} gives it under each of the four learners; the best learner per rung in Table~\ref{tab:ladder} is chosen on out-of-fold error and is descriptive. The remaining eleven include the other economic benchmarks, so the test measures transfer within the battery from companions and academic columns alike; withholding the whole economic block at once is the sharper test we did not run. Because the factors are estimated quantities, we re-estimate them inside the loop by fitting the factor analysis on each training fold alone, projecting the held-out fold through the training-fold Thurstone factor-score weights, and re-orienting by the training-fold mean index. A naive fit-on-all-data pipeline would silently open that leakage path (full procedure in Appendix~\ref{app:lobo}).

Two error metrics then serve two purposes. Root mean squared error (RMSE) on the standardised target is the descriptive metric since it stays on the target's own scale. The test statistic is $\dmse = \mathrm{MSE}(\text{single index}) - \mathrm{MSE}(k \text{ factors})$ per economic target, positive when the multi-factor representation wins, with a 95\% interval from a paired bootstrap of the pooled out-of-fold errors \citep{efron1993} with $B = 2000$.

\subsection{Pre-specified hypotheses}
\label{sec:hyp}

Before analysis, the design fixed four hypotheses with quantitative thresholds set with reference to the figures \citet{kearns2026} reports for the Open LLM Leaderboard. \textbf{H1} (dominance, RQ1) holds that the first factor carries more than 50\% of common variance. \textbf{H2} (date artefact in the plan's term, which we read as association, RQ2) has two parts, (i) that the dominant factor's release-date $\Rsq$ is at least 0.30 and the strongest of the factors, and (ii) that date adjustment lowers the first factor's share by at least fifteen percentage points. \textbf{H3} (structural distinctiveness, RQ3) holds that after date adjustment parallel analysis retains a factor on which the four economic benchmarks load at least 0.40 and exceed their cross-loadings, which is the discriminant criterion of \citet{campbell1959} applied to a benchmark battery. \textbf{H4} (predictive validity, RQ4) holds that the multi-factor representation's $\dmse$ interval excludes zero and a majority of the four economic targets individually favour it. We report every outcome, including the two that fail.

Two factor counts are used in what follows. The primary count is $k = 1$, returned by parallel analysis, the Kaiser rule, and the scree elbow, and it governs H1 to H3; the share statistic in H1 and H2(ii) is computed on the three-factor extraction, since under $k = 1$ it is 100\% by construction (Appendix~\ref{app:derivations}). The second is $k = 3$, an exploratory over-extraction that no criterion selects, which supplies the loading pattern of Section~\ref{sec:h3} and the $k$-factor predictor of Task~2 and is a documented deviation from the plan (Appendix~\ref{app:deviations}). The predictive conclusion does not hinge on it, since sweeping rung (iv) over $k \in \{2,3,4,5\}$ beats the single index at every count (Appendix~\ref{app:robust}).

\section{Results}
\label{sec:results}

The results follow the four hypotheses in order, and the two verdicts that fail are reported as they fell.

\subsection{Dominance of a single factor (H1)}
\label{sec:h1}

Every dimensionality diagnostic points to a dominant capability axis. The first principal component alone accounts for 79.4\% of total score variance, and parallel analysis retains a single factor, since only the first observed eigenvalue of 9.53 exceeds its random-data 95th percentile of 1.79 (Appendix Figure~\ref{fig:scree}). In the three-factor maximum-likelihood extraction the first factor holds 74.5\% of common variance, above the 50\% dominance threshold, so the dominant general factor reported across the psychometric literature (Section~\ref{sec:related}), including the 72\% \citet{kearns2026} reports under a different specification, replicates on a frontier snapshot. H1 is therefore supported, which answers RQ1 in full, and because rank-based and logit-transformed variants only raise the share (Section~\ref{sec:robust}), the Pearson-based headline is the conservative one.

\subsection{Release date and the dominant factor (H2)}
\label{sec:h2}

\begin{figure}[!t]
\centering
\includegraphics[width=0.83\textwidth]{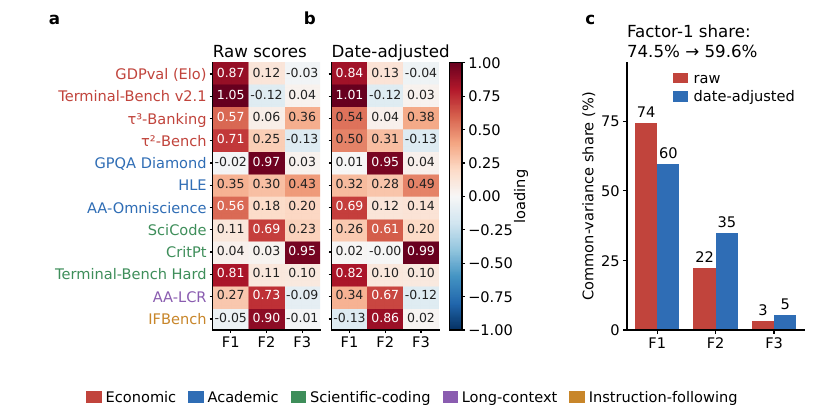}
\caption{\textbf{Factor loadings from the three-factor oblimin EFA.} (a) Raw scores; the economic benchmarks (red) already concentrate on F1. (b) After date adjustment they still concentrate on F1, with a largest cross-loading of 0.38. (c) The first factor's common-variance share falls from 74.5\% to 59.6\%, a 14.9-point drop.}
\label{fig:loadings}
\end{figure}

The factor's scores track model release date with a logistic-fit $\Rsq$ of 0.505 (0.477 under an ordinary-least-squares fit, Appendix~\ref{app:robust}), which is the strongest date association of the three factors. H2(i) is therefore supported, and the leading axis of ``capability'' is closely associated with release date. Adjusting every benchmark for release date and re-fitting then lowers the first factor's common-variance share from 74.5\% to 59.6\% (Figure~\ref{fig:loadings}c), a drop of 14.9 percentage points. That point estimate falls just short of the pre-specified fifteen-point threshold, with a bootstrap 95\% interval of $[-5.3, +32.7]$, so on the primary grid the drop is indistinguishable both from zero and from the threshold. We report H2(ii) as not met on the point estimate without moving the line after seeing the result.

Whether H2(ii) passes depends on the unit of analysis. Under the plan's base-model unit the drop is 24.1 points and clears the threshold; under the configuration-level primary it is 14.9 points and fails; on the 58-model compute-known subsample date alone gives 16.5 points and clears it. Two of the three specifications therefore pass. Promoting the deduplicated grid to primary would pass the threshold on fewer models (89 against 96) and turn the plan's robustness unit into the headline, so we keep the configuration level as primary. The drop clears the threshold on the subsample and the deduplicated grid, which we report as consistent with H2(ii) while the primary verdict stands as not met. RQ2 is therefore answered in part, since the leading axis tracks release date closely (H2(i)) while the pre-specified size of the correction is not met on the primary grid (H2(ii)). Adding training compute does not deepen the correction, since on the compute-known subsample the joint date-plus-compute drop is smaller at 9.3 points. The likely reason is that later models are also larger and the two predictors split their shared variance.

\subsection{Economic distinctiveness (H3)}
\label{sec:h3}

After date adjustment we test whether the economic benchmarks form a distinct factor, which requires that parallel analysis retain a factor and that its loadings concentrate discriminantly on the economic block. On the date-adjusted data parallel analysis retains a single factor, which leaves no block to separate; the separation appears only once three factors are extracted, which the dimensionality rule does not endorse. Holding H3 to the plan as H2(ii) was held to its 15-point line, we report H3 as not supported as specified, which answers RQ3 negatively under the plan.

Under the three-factor over-extraction the economic block does separate, and we report the pattern as exploratory. All four economic benchmarks load at least 0.40 on a single factor, F1, with GDPval at 0.84, Terminal-Bench v2.1 at 1.01, \tauthree{} at 0.54, and \tautwo{} at 0.50, against a largest economic cross-loading of 0.38 (Figure~\ref{fig:loadings}b, tabulated in Appendix~\ref{app:loadings}). The mean absolute F1 loading is 0.72 for the economic benchmarks and 0.32 for the rest. Because the three factors are strongly correlated (0.67 between F1 and F3 and 0.72 between F2 and F3) the separation reads as a secondary refinement of one dominant axis. Bootstrap intervals on the economic loadings are wide (0.10 to 0.78 for \tautwo{} and 0.44 to 0.99 for GDPval), excluding zero but overlapping one another, so the pattern is suggestive and loosely estimated.

\subsection{Predictive validity (H4)}
\label{sec:h4}

\begin{figure}[!t]
\centering
\includegraphics[width=0.85\textwidth]{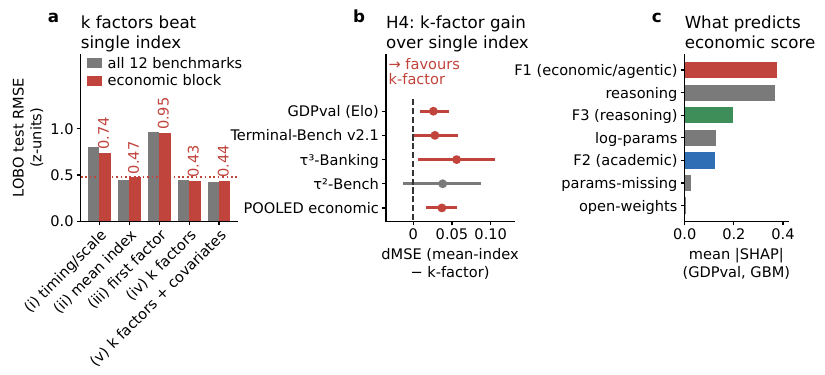}
\caption{\textbf{Predictive validity under the LOBO protocol.} (a) Test RMSE by rung, economic block (red) and all twelve targets (grey). (b) Bootstrap $\dmse$ (single index minus $k$ factors) with 95\% intervals; positive favours the multi-factor representation. (c) SHAP (SHapley Additive exPlanations) importances for the rung-(v) gradient-boosting model of GDPval, the learner that admits exact SHAP values.}
\label{fig:lobo}
\end{figure}

The final test asks whether a multi-factor representation of the other benchmarks predicts a held-out economic benchmark better than their mean. Across the predictor ladder a single mean-score index is already a strong baseline, at economic-block test RMSE 0.474 and $\Rsq$ 0.771; the $k$-factor representation improves on it to RMSE 0.433 and $\Rsq$ 0.808, and adding covariates gives no further gain (Table~\ref{tab:ladder}, Figure~\ref{fig:lobo}a). The first factor alone predicts poorly at RMSE 0.950. Bootstrapping $\dmse$ over the pooled out-of-fold errors gives a pooled improvement of $+0.037$ with a 95\% interval of $[+0.019, +0.055]$, and because that resample includes near-duplicate settings of the same base model, we re-run it on the deduplicated grid of 89 distinct models, where the improvement is $+0.038$ with $[+0.020, +0.056]$ (Appendix~\ref{app:robust}). Three of the four economic benchmarks improve individually, GDPval by $+0.026$, Terminal-Bench v2.1 by $+0.028$, and \tauthree{} by $+0.056$, while \tautwo{}'s interval includes zero (Table~\ref{tab:ladder}, Figure~\ref{fig:lobo}b). H4 is therefore supported under ridge, the best-fitting learner, on the rule of Section~\ref{sec:hyp}; the plan fixed no learner, and Section~\ref{sec:robust} reports the tree-learner reversal.

\begin{table}[!t]
\caption{\textbf{Predictor-ladder cross-validated errors (economic block) and the H4 bootstrap test.} Left, train and test RMSE on standardised targets averaged over the economic targets, out-of-fold $\Rsq$, and the best learner per rung (RF random forest, EN elastic net). Right, $\dmse = \mathrm{MSE}(\text{mean index}) - \mathrm{MSE}(k\ \text{factors})$, $B = 2000$; every interval except \tautwo{}'s excludes zero.}
\label{tab:ladder}
\centering
\footnotesize
\setlength{\tabcolsep}{1.5pt}
\begin{minipage}[t]{0.51\textwidth}
\centering
\begin{tabular}{llrrr}
\toprule
Rung & Learner & Train & Test & $\Rsq$ \\
\midrule
(i) timing / scale & RF & 0.521 & 0.741 & 0.459 \\
(ii) mean index & Ridge & 0.463 & 0.474 & 0.771 \\
(iii) first factor & EN & 0.933 & 0.950 & 0.110 \\
(iv) $k$ factors & Ridge & 0.410 & 0.433 & 0.808 \\
(v) $k$ factors + cov.\ & Ridge & 0.391 & 0.438 & 0.800 \\
\bottomrule
\end{tabular}
\end{minipage}\hfill
\begin{minipage}[t]{0.47\textwidth}
\centering
\begin{tabular}{lrl}
\toprule
Target & $\dmse$ & 95\% CI \\
\midrule
GDPval (Elo) & $+0.026$ & $[+0.010, +0.044]$ \\
Terminal-Bench v2.1 & $+0.028$ & $[+0.001, +0.056]$ \\
\tauthree & $+0.056$ & $[+0.008, +0.103]$ \\
\tautwo & $+0.038$ & $[-0.011, +0.086]$ \\
\textbf{Pooled economic} & $\mathbf{+0.037}$ & $\mathbf{[+0.019, +0.055]}$ \\
\bottomrule
\end{tabular}
\end{minipage}
\end{table}

The 0.037 gain sits on top of a single index that already explains 77\% of economic-block variance. It shows that the battery carries structure a single index misses. Whether an economic benchmark adds information the others lack is a separate question, which needs the economic block withheld as a whole or an external criterion. The one exception, \tautwo{}, carries the widest interval, and it is also the economic benchmark with the lowest F1 loading (0.50) and the one that branches alone in the benchmark clustering (Appendix~\ref{app:clustering}), so its null is consistent with its structural separation from the other three.

\subsection{Robustness}
\label{sec:robust}

The headline structure is stable under perturbations to the unit of analysis, the correlation measure, the adjustment covariates, and the factor count (Appendices~\ref{app:deviations} and~\ref{app:robust}). Collapsing the 421 configurations to one row per base model raises the first-factor share to 90.2\% and the date-adjustment drop to 24.1 points, and rank-based and logit-transformed correlations raise the share to 94.5\% and 91.8\%, so H1 is no metric artefact \citep{schaeffer2023}. The H4 verdict depends on neither the regularisation strength nor the factor count, and it does depend on learner family. Ridge and elastic net favour the multi-factor representation on all four targets, and the tree learners, which fit the economic block worse on both rungs, reverse it (C8). The $k$-factor advantage also extends beyond the economic block, averaging 0.048 ridge against ridge on the eight other targets (positive on seven) against 0.037 on the economic four, with intervals computed for the economic block only.

\section{Discussion}
\label{sec:discussion}

Taken together, the two tasks answer the four research questions, and the answers do not line up. On RQ1 the battery is governed by one dominant axis holding 74.5\% of common variance, consistent with prior reports; on RQ2 that axis tracks release date ($\Rsq = 0.505$), and date adjustment lowers its share by about a fifth on the point estimate (74.5\% to 59.6\%), with an interval that spans zero on the primary grid. On RQ3 the economic benchmarks form no separate factor under the pre-specified rule and separate only under an over-extraction, yet on RQ4 a multi-factor representation still predicts held-out economic scores better than a single general index, on three of four benchmarks and on the deduplicated grid, and it predicts the other benchmarks better too, all under the linear learners that fit best. The structural and predictive tests disagree, and the disagreement is the finding. A retention rule asks whether the economic benchmarks add enough shared variance to move an eigenvalue past a random-data threshold; a held-out prediction asks whether a richer representation of the battery predicts them better than a general index does. The second can be answered yes while the first is answered no, and two features of the study produce that gap without any flaw in either test. The tests set different bars, since a retention rule needs an eigenvalue above what random data of the same shape produce, where at $n = 96$ that line sits at 1.57 against an observed second eigenvalue of 0.91, while an out-of-sample gain of a few hundredths of a standardised unit clears a bootstrap interval. The sample is small for the structural test and adequate for the predictive one, which is why the bootstrap interval on the date drop spans zero while the interval on $\dmse$ does not. A single index reproduces most of what distinguishes models, 77\% of economic-block variance out of sample, and that agreement does not license the claim that the economic benchmarks add nothing at the margin. A price index built from benchmark scores shows the same shape, since removing contamination-flagged benchmarks leaves its model ranking at a rank correlation of 0.998 while the index moves by 0.49 log points a year \citep{zhu2026price}. Construct validity should therefore be assessed with a predictive test beside the structural one, and a validity claim for a benchmark should say which test it passed.

For leaderboard use, a single intelligence index captures most of what distinguishes models, and the structure it misses carries predictive information for every column, which justifies reporting the benchmarks separately, the economic ones among them. A leaderboard ranking models released months apart is ranking them largely along an axis that tracks release date. A comparison within one release window holds that trend fixed, as a buyer choosing between contemporaneous models needs.

A second contribution follows for benchmark builders in the form of a two-test protocol, since a benchmark's value for distinguishing contemporaneous models depends on the variance it adds once the shared date trend is removed. A builder who wants to show that a new suite measures something distinct can apply the two tests directly with what a leaderboard already publishes, per-model scores on the new benchmark and the existing battery and their release dates. Step one, the structural test, regresses every benchmark on release date, fits an oblique factor model to the residuals, and checks whether the new benchmark's loading separates from the general factor. Step two, the predictive test, holds the new benchmark out, predicts it from a $k$-factor representation of the others with factors re-estimated in every fold and a regularised linear learner, reporting tree learners beside it, and checks whether that beats a single mean-score index with a bootstrap interval excluding zero. A benchmark that passes both adds a construct, one that passes only the second shares structure with the battery beyond the general index without forming a factor of its own, and one that passes neither is re-measuring the progress trend. Step two presupposes companions in the battery, and a benchmark for a construct the battery does not yet represent needs an external criterion. An operator can run both tests on every new benchmark and publish the two statistics beside the score, and the accompanying code does so from a single score matrix (\url{https://github.com/louisyzhu/frontier-ai-economic-validity}).

Four limitations bound the claims, in decreasing order of consequence. First, the economic factor comes from an exploratory over-extraction, so it stands as a hypothesis for a confirmatory factor model on an independent snapshot. Second, the analysis is observational. Release date stands in for everything that arrives with newer models, so the component the adjustment removes contains real improvement along with any calendar effect, and the counterfactual that would separate them, an older and a newer model evaluated under the same conditions at the same time, is one a leaderboard cannot supply. Third, coverage is uneven, so the economic results carry wider uncertainty, and fourth, the results describe one leaderboard on one date. Both tests are internal to the leaderboard, so they leave open whether scores that pass them predict performance on naturalistic professional tasks with tools and interaction, assessed independently of the benchmarks and with any automated grader's validity measured \citep{zhu2026judge}.

Economic benchmarks therefore measure neither one capability nor many. Most of what they record is a general factor that tracks release date, and under the linear learners the structure one index misses predicts held-out scores without forming a factor of its own. Hypotheses and thresholds were fixed in advance, Appendix~\ref{app:deviations} lists every departure, and every verdict stands as it fell.

\section*{Acknowledgements}

We thank Marcos Barreto and Thomas Robinson for comments on successive drafts of this paper.

\paragraph{Generative AI use.} Generative AI tools assisted with tasks ancillary to the research, namely reference verification and formatting, appendix and figure preparation, auditing and debugging of the released code, and consistency checks of numbers and cross-references against the archived outputs. They were not used to generate research questions or hypotheses, to design or run the analysis, or to draw its conclusions. The research and its text are our own, every AI-assisted output was reviewed and revised by us, and we take full responsibility for the final content.

\paragraph{Reproducibility.} Every number in the paper traces to two SHA-256-pinned snapshots described in Appendix~\ref{app:provenance}. The analysis code is released as benchprobe v0.1.0 \citep{benchprobe2026}, extracted from the archived analysis notebook and released under the MIT licence, with the derived tables under CC BY 4.0 and the leaderboard snapshot redistributed under the operator's public-page terms. One command recomputes the results reported here from the pinned 6 July 2026 snapshot, and every table the package emits states its reproducibility tier and its deviation from the archived value. Recomputed outputs are produced again from the pinned inputs and agree with the archived table within a stated tolerance; they cover the structure results (the KMO of 0.933, the first-factor share of 74.5\%, the residualised share of 59.6\% and its 14.9-point drop, and the residualised oblimin loadings) and the leave-one-benchmark-out point estimates, including the pooled economic $\dmse$ of $+0.037$ and the four-learner ladder, which reproduces the archived rung summary exactly, the summary's per-rung error being the mean of the per-target RMSEs. Statistically reproduced outputs are produced again under a random stream the archive does not record and agree within three Monte-Carlo standard deviations; they cover the bootstrap interval $[+0.019, +0.055]$ and the $k$-sweep. Regenerated outputs are copied from the archive and re-rendered without recomputation; they cover the clustering table, which lies outside the library's scope, and the logit first-factor share, the calibration constants, and the BIC row, whose computations do not appear in the archived notebook. Two quantities were reconstructed because the archive ships no code for them, and a reader should weigh them accordingly. The four-parameter logistic behind the date $\Rsq$ of 0.505 was recovered by fitting candidate forms until one reproduced the archived value; the analysis plan specifies a logistic fit without fixing its parameterisation, so the recovered form is consistent with the plan. The pooling rule behind the ladder table was inferred as the mean of per-target RMSEs, which reproduces all ten archived rows, whereas the notebook's own helper pools by root mean square and reproduces none. The analysis plan, fixed before the data were captured, is deposited at \url{https://doi.org/10.17605/OSF.IO/VD34J}, and the deviations from it are documented in Appendix~\ref{app:deviations}.

\bibliographystyle{plainnat}
\bibliography{refs}

\newpage
\appendix
\section*{Appendix}
The appendix collects the material that supports the main text without interrupting its argument. Each section opens with the main-text section it supports, and every figure and table is placed with the section that discusses it.

\section{Abbreviations and notation}
\label{app:abbrev}

\begin{table}[H]
\caption{Abbreviations and symbols, in order of first appearance in the main text.}
\label{tab:abbrev}
\centering
\footnotesize
\setlength{\tabcolsep}{3pt}
\renewcommand{\arraystretch}{0.92}
\begin{tabular}{ll@{\hspace{1.2em}}ll}
\toprule
GPQA & Graduate-Level Google-Proof Q\&A & LOBO & leave-one-benchmark-out \\
HLE & Humanity's Last Exam & RMSE & root mean squared error \\
LLM & large language model & MSE & mean squared error \\
RQ & research question & $\dmse$ & $\mathrm{MSE}(\text{single index}) - \mathrm{MSE}(k\text{ factors})$ \\
H1--H4 & pre-specified hypotheses & CI & confidence interval \\
KMO & Kaiser--Meyer--Olkin sampling adequacy & OLS & ordinary least squares \\
EFA & exploratory factor analysis & BIC & Bayesian information criterion \\
PCA & principal-component analysis & SHAP & SHapley Additive exPlanations \\
ML & maximum likelihood & RF, EN & random forest, elastic net \\
$\Rsq$ & coefficient of determination & GBM & gradient-boosting machine \\
$\rho$ & Spearman rank correlation & R1, C1--C8 & planned and delivered robustness checks \\
$k$ & number of retained factors & AA & Artificial Analysis (leaderboard operator) \\
$\Lambda, \Phi, \Psi$ & loadings, factor covariance, uniquenesses & $W$ & Thurstone factor-score weights \\
\bottomrule
\end{tabular}
\end{table}

\section{Leave-one-benchmark-out procedure}
\label{app:lobo}
\emph{Supports Section~\ref{sec:task2}, which describes the protocol in words.} Figure~\ref{fig:lobo-proc} states the complete procedure, including the two safeguards on which the predictive test depends. First, the factor model is re-fitted inside every outer training fold, so the held-out models never influence the loadings used to score them. Second, hyperparameters are tuned in an inner loop nested within the outer split \citep{varma2006}, and the H4 contrast is reported ridge against ridge; the plan does not fix that learner, and ridge is also the lowest-error learner on both rungs, so check C8 in Appendix~\ref{app:robust} gives the contrast under all four. The best learner per rung in Table~\ref{tab:ladder} is chosen on pooled out-of-fold error, as the procedure below shows, so that column is descriptive and its errors carry the optimism of that choice.

\begin{figure}[!htb]
\centering
\small
\fbox{\begin{minipage}{0.95\textwidth}
\textbf{Procedure} \quad Leave-one-benchmark-out predictive validity with nested cross-validation\\[2pt]
\textbf{Input:} raw score matrix $X$ (models $\times$ benchmarks), economic targets $T$, learners $L$, rungs $R$.\\
\textbf{for} each target $t \in T$ \textbf{do}\\
\hspace*{1em} partition models into 5 outer folds\\
\hspace*{1em} \textbf{for} each outer fold (train, test) \textbf{do}\\
\hspace*{2em} drop column $t$; fit standardisation and EFA on train only; form Thurstone weights $W = R^{-1}\Lambda$\\
\hspace*{2em} project test models through $W$; orient factors by train mean index\\
\hspace*{2em} \textbf{for} each rung $r \in R$ and learner $\ell \in L$ \textbf{do}\\
\hspace*{3em} tune $\ell$ by inner 5-fold grid search on train; predict test, store out-of-fold error\\
\hspace*{1em} select best learner per rung by pooled out-of-fold RMSE (the H4 contrast is ridge against ridge and does not depend on this choice)\\
compute $\dmse = \mathrm{MSE}(\text{rung ii}) - \mathrm{MSE}(\text{rung iv})$; bootstrap $B = 2000$ for 95\% CI\\
\textbf{Output:} per-target and pooled $\dmse$ with confidence intervals (H4).
\end{minipage}}
\caption{The leave-one-benchmark-out procedure with leakage-safe factor projection and nested tuning.}
\label{fig:lobo-proc}
\end{figure}

\section{Positionality, directionality, and threats to validity}
\label{app:positionality}
\emph{Supports Section~\ref{sec:hyp}.} A factor solution is identified only up to sign, so we state the direction of every factor explicitly. We orient each factor to correlate positively with the model's mean standardised score. ``Higher'' then always means ``more capable'', and every loading in Table~\ref{tab:loadings} reads in the same direction. The choice is not cosmetic, since an un-oriented solution could report the same economic factor with flipped signs and invite the opposite interpretation. The direction of the date adjustment is equally deliberate. We regress scores on release date and analyse the residuals, which measures how much benchmark co-movement survives once the release date is known. We do not run the reverse regression, as we make no claim that the general factor causes the calendar. Finally, we fix the sign of the predictive contrast in advance, so that $\dmse$ is the single-index error minus the $k$-factor error, a positive value favours the richer representation, and H4 requires this positive interval to exclude zero.

Two threats to validity follow from the study's vantage point. First, we observe only the models that vendors chose to submit to a public leaderboard. The sample is therefore a convenience sample of frontier systems, and the claims generalise to models on this leaderboard alone. Second, the economic benchmarks are newer and sparser, so any economic-specific structure is estimated from fewer observations and is more sensitive to the particular models that carry those scores. The wide bootstrap intervals in Table~\ref{tab:ladder} quantify that limitation.

\section{Mathematical basis of the two tasks}
\label{app:derivations}
\emph{Supports Sections~\ref{sec:task1} and~\ref{sec:task2}.} Each subsection states a component of the method, derives the quantity the main text reports, and gives the reason the component was chosen over its nearest alternative.

\paragraph{The common-factor model and the dominance statistic.} Let $\mathbf{z} \in \mathbb{R}^{p}$ be the vector of $p$ per-benchmark standardised scores for a model. The common-factor model writes
\[
\mathbf{z} = \Lambda \mathbf{f} + \boldsymbol{\varepsilon}, \qquad \Sigma = \Lambda \Phi \Lambda^{\top} + \Psi,
\]
with $\Lambda \in \mathbb{R}^{p \times k}$ the loading matrix, $\mathbf{f} \in \mathbb{R}^{k}$ the latent factors with $\mathrm{Cov}(\mathbf{f}) = \Phi$, and $\boldsymbol{\varepsilon}$ benchmark-specific uniquenesses with diagonal $\mathrm{Cov}(\boldsymbol{\varepsilon}) = \Psi$ and $\mathrm{Cov}(\mathbf{f}, \boldsymbol{\varepsilon}) = \mathbf{0}$. The decomposition of $\Sigma$ into a common part $\Lambda\Phi\Lambda^{\top}$ and a unique part $\Psi$ is the reason we use factor analysis in preference to principal components for a validity question. A principal component maximises total variance and so absorbs benchmark-specific noise along with shared capability, whereas the common-factor model attributes to the factors only the variance that benchmarks share \citep{thurstone1947}. Parameters are estimated by maximum likelihood under multivariate normality. The dominance statistic for H1 is computed on the unrotated solution, where $\Phi = I$ and each factor's contribution to common variance is its sum of squared loadings, the $m$-th diagonal entry of $\Lambda^{\top}\Lambda$,
\[
s_{1} = \frac{\sum_{j=1}^{p} \lambda_{j1}^{2}}{\sum_{m=1}^{k} \sum_{j=1}^{p} \lambda_{jm}^{2}} = \frac{(\Lambda^{\top}\Lambda)_{11}}{\mathrm{tr}(\Lambda^{\top}\Lambda)}.
\]
On the complete-case grid the three factors of the exploratory extraction carry sums of squared loadings 7.79, 2.34, and 0.32, so $s_{1} = 7.79/10.45 = 74.5\%$; the statistic is defined on that extraction because under $k = 1$ it equals one by construction, and the first principal component's 79.4\% of total variance is the count-free companion figure. This is a factor-level share, distinct from a single benchmark's communality $\sum_{m} \lambda_{jm}^{2}$, a diagonal entry of $\Lambda\Lambda^{\top}$.

\begin{figure}[!htb]
\centering
\includegraphics[width=\textwidth]{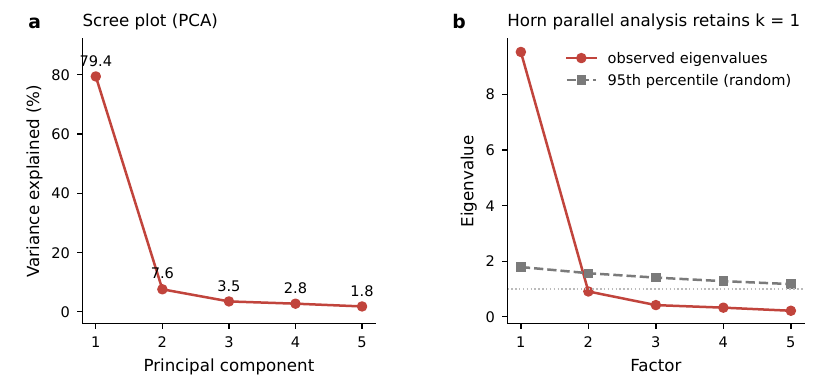}
\caption{\textbf{Dimensionality of the battery.} (a) The scree plot of principal-component variance shares drops sharply after the first component (79.4\%). (b) Horn parallel analysis retains a single factor, as only the first observed eigenvalue (9.53) exceeds its random-data 95th percentile (1.79), while the second (0.91) already falls below (1.57).}
\label{fig:scree}
\end{figure}

\paragraph{Dimensionality by parallel analysis.} Let $\ell_{1} \geq \dots \geq \ell_{p}$ be the eigenvalues of the sample correlation matrix $R$, and let $\ell_{m}^{(0.95)}$ be the 95th percentile of the $m$-th eigenvalue over correlation matrices of independent standard-normal data of the same dimensions $n \times p$. Parallel analysis retains factor $m$ when $\ell_{m} > \ell_{m}^{(0.95)}$ \citep{horn1965}. The rule asks whether a factor explains more than chance sampling structure would, so it is more conservative than the Kaiser rule $\ell_{m} > 1$ \citep{kaiser1960} whenever finite-sample noise pushes the random-data eigenvalues above one, as it does here ($\ell^{(0.95)}_{2} = 1.57$); on this battery both rules retain one factor. Information criteria such as the BIC \citep{schwarz1978} compare likelihoods of nested models and reward any factor that improves fit, which on a near-rank-one correlation matrix means retaining factors that carry very little common variance; the BIC's preference for $k = 4$ here (check C7, Appendix~\ref{app:robust}) is an instance. On our grid $\ell_{1} = 9.53$ against $\ell_{1}^{(0.95)} = 1.79$, and $\ell_{2} = 0.91$ against $\ell_{2}^{(0.95)} = 1.57$, so one factor is retained (Figure~\ref{fig:scree}).

\paragraph{Oblique rotation.} The maximum-likelihood solution is identified only up to rotation. Direct oblimin \citep{jennrich1966} chooses the rotation that minimises
\[
Q(\Lambda) = \sum_{m < l} \Bigl( \sum_{j=1}^{p} \lambda_{jm}^{2} \lambda_{jl}^{2} - \frac{\gamma}{p} \sum_{j=1}^{p} \lambda_{jm}^{2} \sum_{j=1}^{p} \lambda_{jl}^{2} \Bigr),
\]
with $\gamma = 0$ giving the quartimin criterion we use, while allowing $\Phi \neq I$. An orthogonal rotation such as varimax forces $\Phi = I$; on data with a strong general factor this distributes shared variance across artificially independent axes, so that no single factor can carry the general variance and the dominance we set out to measure is obscured by construction. The date-adjusted factors correlate at 0.67 and 0.72 (Section~\ref{sec:h3}), which confirms that the oblique choice was the right one.

\paragraph{Factor scores and the leakage-safe projection.} Regression (Thurstone) factor scores are the linear predictor $\hat{\mathbf{f}} = W^{\top}\mathbf{z}$ that minimises $\mathbb{E}\lVert \mathbf{f} - W^{\top}\mathbf{z} \rVert^{2}$. Setting the derivative to zero gives $W = \Sigma^{-1}\mathrm{Cov}(\mathbf{z}, \mathbf{f}) = R^{-1}\Lambda\Phi$, where $R$ is the sample correlation matrix, $\Lambda$ the pattern loadings, and $\Phi$ the inter-factor correlation matrix of the sorted solution, so that $W = R^{-1}\Lambda$ when the factors are orthogonal \citep{thurstone1947}. The structural factor scores behind H2 and the model clustering use the oblique form. The prediction task uses the orthogonal-form weights $W = R^{-1}\Lambda$ built from the pattern loadings; there $W$ is estimated on the training fold only and applied to the held-out fold, the leakage-safe projection of Section~\ref{sec:task2}. The two score sets are an invertible reparameterisation of one another, $\hat{\mathbf{f}}_{\mathrm{oblique}} = \Phi\,\hat{\mathbf{f}}_{\mathrm{pattern}}$, so an unpenalised regression on either gives identical fitted values; the ridge learners of Task~2 are not invariant to the reparameterisation, and the reported ladder uses the pattern form throughout. $W$ is a function of the data, so if it were estimated on all $n$ models, every held-out model's score would have contributed to the weights used to predict it, and the predictive comparison in Task~2 would be contaminated by exactly the information it is meant to withhold.

\paragraph{Date adjustment as removal of a rank-one component.} Each standardised benchmark $z_{j}$ is regressed on release date $d$ (and, in a subsample, additionally on log training compute),
\[
z_{j} = \beta_{0j} + \beta_{1j}\, d + r_{j},
\]
and the factor model is re-fitted on the re-standardised residuals $r_{j}$. The regression is linear in $d$, so the adjustment removes the linear component of the date trend; residualising on the logistic form that fits the date association best is a check we did not run. Writing $\mathbf{b} = (\beta_{11}, \dots, \beta_{1p})^{\top}$, the covariance of the residual vector is $\Sigma_{r} = \Sigma - \sigma_{d}^{2}\, \mathbf{b}\mathbf{b}^{\top}$, so the adjustment removes from the benchmark covariance exactly the rank-one component that lies along the date vector. If the general factor were nothing but the shared time trend, $\Lambda_{1}$ would be proportional to $\mathbf{b}$ and the first factor's share would collapse after adjustment; if it were unrelated to date, the share would be unchanged. The observed change, $74.5\% \rightarrow 59.6\%$, is the H2(ii) statistic and measures where the battery sits between those two cases. The design mirrors the scaling-law residualisation of \citet{kearns2026}, with release date in place of parameter count as the variable adjusted for.

\paragraph{Nested leave-one-benchmark-out estimator.} For an economic target $t$, the out-of-fold error of rung $r$ is
\[
\mathrm{MSE}_{r} = \frac{1}{n} \sum_{i=1}^{n} \bigl( z_{it} - \hat{z}_{it}^{(-\kappa(i))} \bigr)^{2},
\]
where $\hat{z}_{it}^{(-\kappa(i))}$ is the prediction for model $i$ from a learner trained on the outer folds excluding the fold $\kappa(i)$ that contains $i$, with hyperparameters chosen by an inner five-fold grid search on those training folds alone. Selecting hyperparameters on the same folds that supply the reported error biases the estimate optimistically, and the bias grows with the size of the search space \citep{varma2006}; nesting removes it at the cost of a five-fold increase in computation. The predictive-validity statistic contrasts the single-index rung against the $k$-factor rung, $\dmse = \mathrm{MSE}_{\mathrm{ii}} - \mathrm{MSE}_{\mathrm{iv}}$.

\paragraph{Paired bootstrap for $\dmse$.} Let $e_{i}^{\mathrm{ii}}$ and $e_{i}^{\mathrm{iv}}$ be model $i$'s out-of-fold squared errors under the two rungs. A 95\% interval for $\dmse$ is obtained by resampling models with replacement, recomputing $\frac{1}{n}\sum_{i}(e_{i}^{\mathrm{ii}} - e_{i}^{\mathrm{iv}})$ on each of $B = 2000$ resamples, and taking the 2.5th and 97.5th percentiles \citep{efron1993}. Resampling the paired differences matters because the two rungs are evaluated on the same models and their errors are strongly correlated, whereas resampling the two error vectors separately would discard that pairing; the paired design removes the between-model variance that the two rungs share, so the interval reflects only the variance of the contrast. H4 is supported when this interval excludes zero. Because the resampling unit is a model, near-duplicate configurations of one base model would be treated as independent draws; the deduplicated re-run in Appendix~\ref{app:robust}, whose resampling unit is a distinct base model, is the check on that assumption.

\section{Date-adjusted three-factor loadings}
\label{app:loadings}
\emph{Supports Section~\ref{sec:h3}.} Table~\ref{tab:loadings} tabulates the loadings shown as a heatmap in Figure~\ref{fig:loadings}b. F1 is an agentic and work-realistic factor, F2 an academic-knowledge factor, and F3 a physics and hard-reasoning factor; all four economic benchmarks load highest on F1.

\begin{table}[!htb]
\caption{\textbf{Date-adjusted three-factor oblimin loadings}, economic benchmarks in bold.}
\label{tab:loadings}
\centering
\small
\renewcommand{\arraystretch}{0.92}
\begin{tabular}{lrrr}
\toprule
Benchmark & F1 & F2 & F3 \\
\midrule
\textbf{GDPval (Elo)} & \textbf{0.84} & 0.13 & $-0.04$ \\
\textbf{Terminal-Bench v2.1} & \textbf{1.01} & $-0.12$ & 0.03 \\
\textbf{\tauthree} & \textbf{0.54} & 0.04 & 0.38 \\
\textbf{\tautwo} & \textbf{0.50} & 0.31 & $-0.13$ \\
GPQA Diamond & 0.01 & 0.95 & 0.04 \\
HLE & 0.32 & 0.28 & 0.49 \\
AA-Omniscience & 0.69 & 0.12 & 0.14 \\
SciCode & 0.26 & 0.61 & 0.20 \\
CritPt & 0.02 & $-0.00$ & 0.99 \\
Terminal-Bench Hard & 0.82 & 0.10 & 0.10 \\
AA-LCR & 0.34 & 0.67 & $-0.12$ \\
IFBench & $-0.13$ & 0.86 & 0.02 \\
\bottomrule
\end{tabular}
\end{table}

\section{Benchmark battery}
\label{app:battery}
\emph{Supports Section~\ref{sec:data}.} Table~\ref{tab:battery} describes every candidate benchmark, the construct it targets, the type of score on which the leaderboard reports it, its coverage before and after the inclusion rule, and the role the rule assigned, with the source paper for each listed beneath. The economic benchmarks differ from the academic ones in kind as well as in coverage. They score multi-step, tool-using, or human-preference-judged task completion on work-like tasks, whereas the academic benchmarks score single-response accuracy on knowledge or reasoning items. Three of the twelve (AA-Omniscience, AA-LCR, and Terminal-Bench Hard) are constructed or subset by the leaderboard operator itself; all twelve are run by that operator under a single harness \citep{aa2026}.

\begin{table}[!htb]
\caption{\textbf{Candidate benchmarks by capability block}, with the construct each targets, the type of score reported, the number of models scored on the raw 548-configuration snapshot and among the 421 retained, and the role assigned by the inclusion rule. Twelve benchmarks are primary; MMMU-Pro is sensitivity-only for sparse and multimodal coverage; APEX-Agents falls below the sixty-model floor. Higher is better throughout; sources are listed below the table.}
\label{tab:battery}
\centering
\footnotesize
\setlength{\tabcolsep}{3pt}
\resizebox{\textwidth}{!}{%
\begin{tabular}{lllrrl}
\toprule
Benchmark & Construct measured & Score type & Raw & Retained & Role \\
\midrule
\multicolumn{6}{l}{\emph{Economic}} \\
GDPval (Elo) & Paid professional task quality & pairwise-judged Elo & 117 & 112 & primary \\
Terminal-Bench v2.1 & Terminal and agentic task completion & pass rate & 121 & 121 & primary \\
\tauthree & Multi-step banking workflows & task success & 112 & 112 & primary \\
\tautwo & Tool-use agentic tasks & task success & 428 & 413 & primary \\
APEX-Agents & Agentic task completion & task success & 26 & -- & dropped (sparse) \\
\multicolumn{6}{l}{\emph{Academic}} \\
GPQA Diamond & Graduate-level science QA & accuracy & 513 & 421 & primary \\
HLE & Frontier academic knowledge & accuracy & 509 & 421 & primary \\
AA-Omniscience & Broad knowledge, hallucination-penalised & penalised score & 418 & 418 & primary \\
MMMU-Pro & Multimodal understanding & accuracy & 204 & -- & sensitivity only \\
\multicolumn{6}{l}{\emph{Scientific coding}} \\
SciCode & Research-level scientific coding & pass rate & 507 & 421 & primary \\
CritPt & Frontier physics problem solving & accuracy & 422 & 419 & primary \\
Terminal-Bench Hard & Hard terminal coding tasks & pass rate & 421 & 413 & primary \\
\multicolumn{6}{l}{\emph{Long-context / instruction}} \\
AA-LCR & Long-context reasoning & accuracy & 441 & 420 & primary \\
IFBench & Instruction following & accuracy & 437 & 417 & primary \\
\bottomrule
\end{tabular}}
\end{table}

\noindent\emph{Sources.} GDPval \citep{patwardhan2025gdpval}; Terminal-Bench v2.1 \citep{merrill2026terminalbench}; \tauthree{} \citep{shi2026tauknowledge}; \tautwo{} \citep{barres2025tau2}; GPQA Diamond \citep{rein2023gpqa}; HLE \citep{phan2025hle}; SciCode \citep{tian2024scicode}; CritPt \citep{zhu2025critpt}; IFBench \citep{pyatkin2025ifbench}; AA-Omniscience, AA-LCR, and Terminal-Bench Hard are constructed or subset by the leaderboard operator \citep{aa2026}.

GDPval's Elo rating comes from pairwise, head-to-head judgements of model outputs on economically valuable tasks, so a higher rating means more often preferred on a scale with no fixed maximum. The $\tau$-Bench family \citep{yao2024taubench,barres2025tau2} scores an agent on whether the final database state and required confirmations match a policy-compliant goal after a multi-turn conversation with a simulated user; \tauthree{} is the banking-knowledge domain of its third generation, in which agents must complete policy-compliant state changes over a large document base. The leaderboard labels its Terminal-Bench column v2.1; the cited paper presents the suite's 2.0 release. MMMU-Pro \citep{yue2025mmmupro} is demoted to a sensitivity-only role and APEX-Agents \citep{vidgen2026apex} is dropped for sparsity (Appendix~\ref{app:coverage}).

\section{Full coverage audit}
\label{app:coverage}
\begin{figure}[p]
\centering
\includegraphics[width=0.88\textwidth]{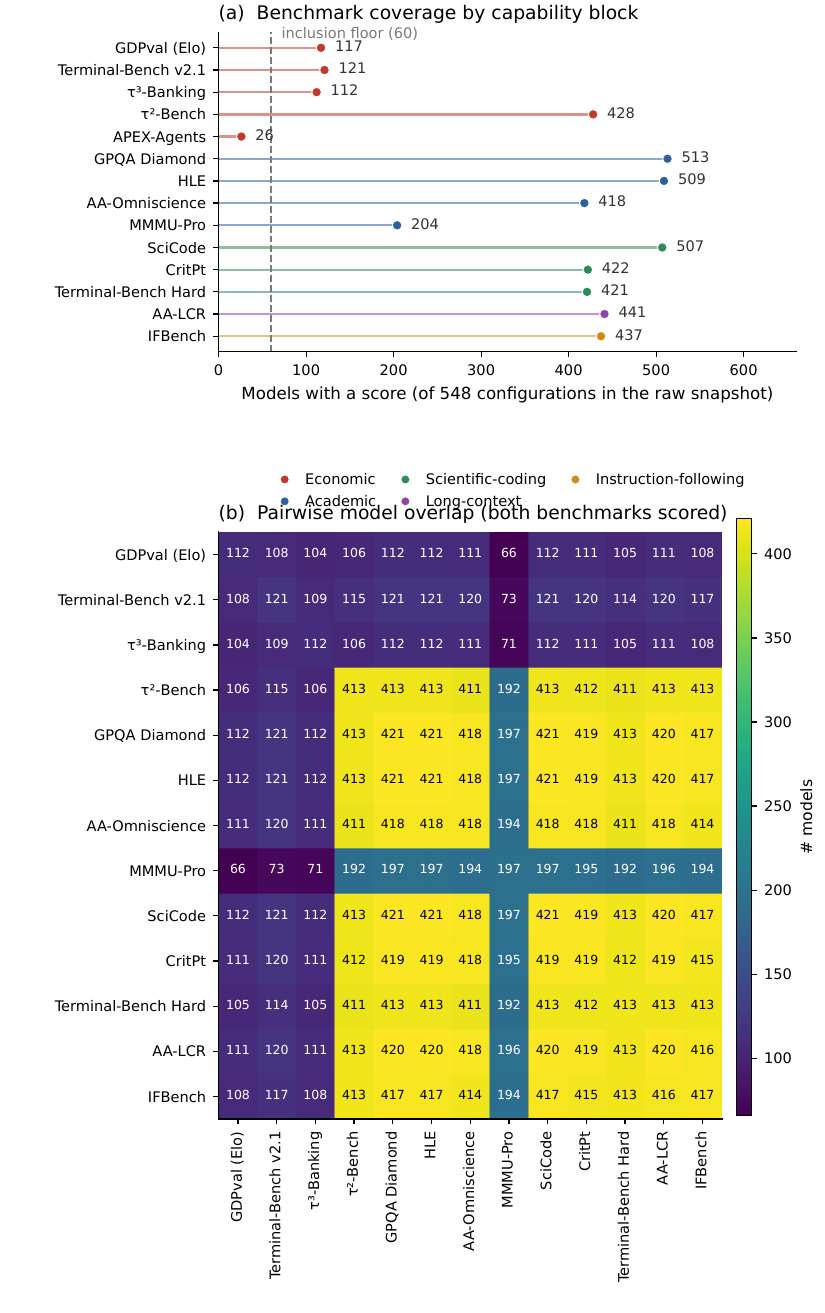}
\caption{\textbf{Coverage audit behind the inclusion rule.} (a) Number of models carrying a score on each candidate benchmark on the raw 548-configuration snapshot, coloured by capability block, with the sixty-model inclusion floor marked. Three economic benchmarks (Terminal-Bench v2.1, GDPval, and \tauthree) are scored on only 112 to 121 models, while \tautwo{} and the academic core are each scored on four hundred or more; APEX-Agents, with 26 models, falls below the floor. (b) Pairwise counts of models scored on both benchmarks in a pair, which set the effective sample size for every correlation in the study and are the evidence for the two-grid design.}
\label{fig:coverage}
\end{figure}

\emph{Supports the inclusion rule of Section~\ref{sec:data}.} Table~\ref{tab:battery} records every candidate benchmark's coverage on the raw snapshot and the decision applied to it, and Figure~\ref{fig:coverage} shows that coverage together with the pairwise overlaps that set the effective sample size for every correlation in the study. The sixty-model floor is the smallest count at which the sparsest retained benchmark still overlaps every other benchmark on more than one hundred models (Figure~\ref{fig:coverage}b), which keeps each pairwise correlation estimable with a standard error below about 0.1.

\section{Exploratory data analysis}
\label{app:eda}
\begin{figure}[!tp]
\centering
\includegraphics[width=\textwidth]{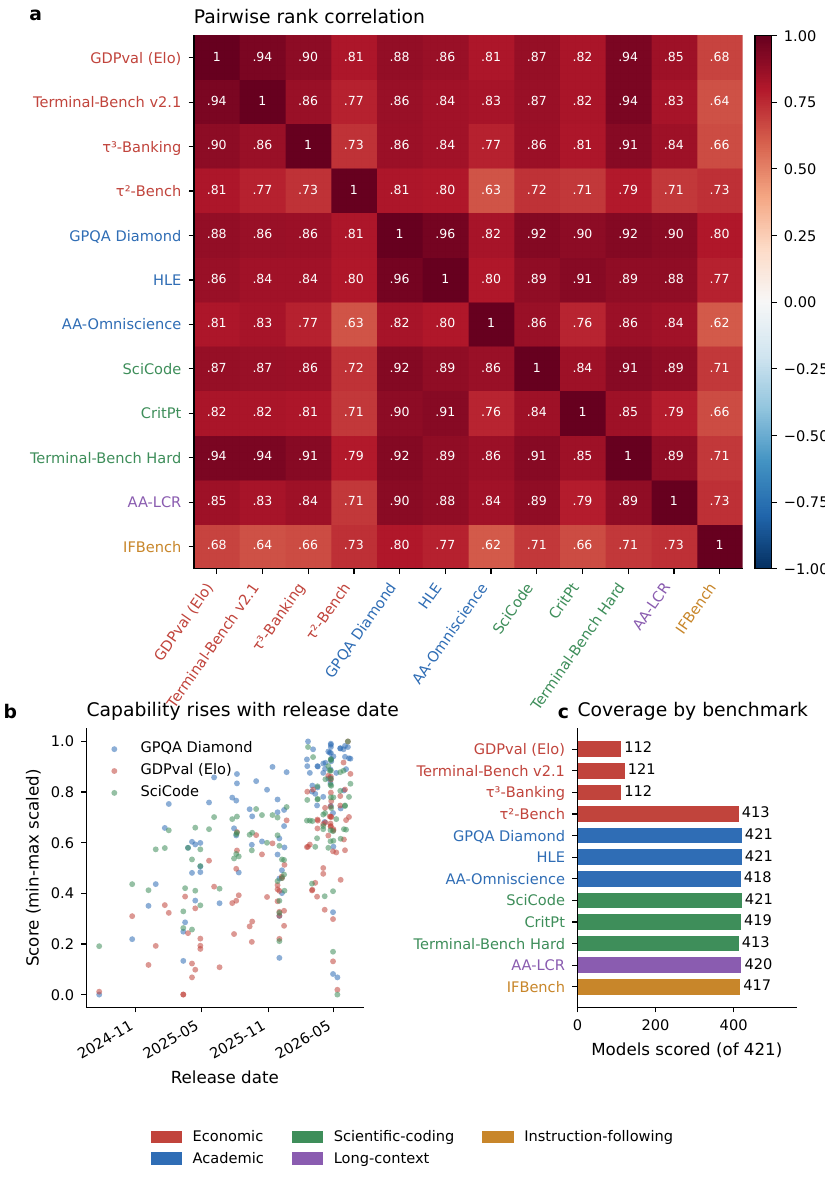}
\caption{\textbf{Exploratory data analysis of the twelve-benchmark battery on the complete-case grid ($n = 96$).} (a) Pairwise Spearman rank correlations, with benchmark labels coloured by capability block; almost every pair is strongly positive. (b) Per-benchmark scores (min--max scaled) against model release date, showing capability rising steeply with time. (c) Coverage per benchmark across the 421 retained models, with the four economic benchmarks (red) scored on far fewer models than the academic core.}
\label{fig:eda}
\end{figure}

\emph{Supports Section~\ref{sec:data}.} We examined coverage and missingness, correlation structure, and temporal patterns before fitting any model (Figure~\ref{fig:eda}). All standardisation and model fitting uses the training portion of each split only, so these views characterise the data without leaking test information into the models.

\begin{figure}[!tp]
\centering
\includegraphics[width=\textwidth]{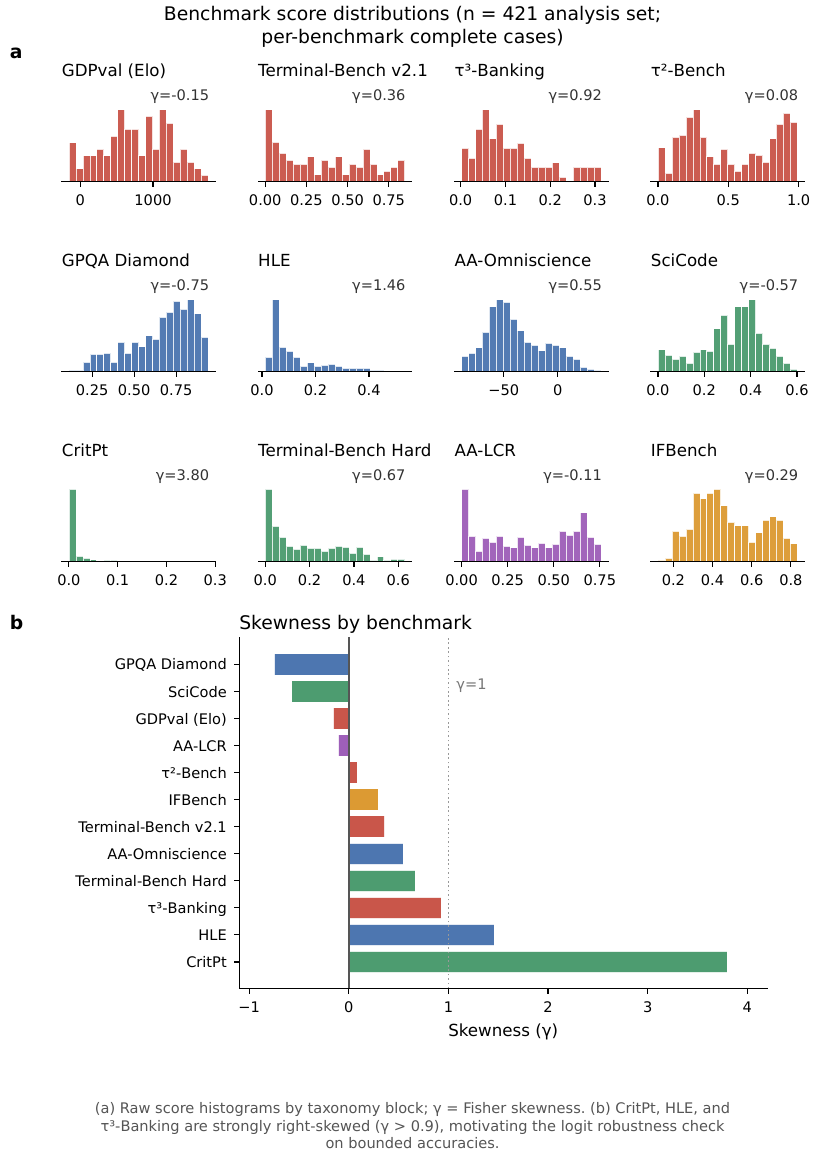}
\caption{\textbf{Univariate benchmark score distributions.} (a) Raw score histograms for the twelve benchmarks, coloured by capability block; $\gamma$ is the Fisher skewness. (b) Skewness ordered across benchmarks. Most benchmarks are bounded accuracies in $[0,1]$, and several are strongly right-skewed (CritPt $\gamma = 3.8$, HLE $\gamma = 1.5$, \tauthree{} $\gamma = 0.9$), with scores concentrated near the floor. GDPval (Elo) and AA-Omniscience are on unbounded scales and are closer to symmetric.}
\label{fig:dist}
\end{figure}

The battery is highly collinear before any modelling. The mean off-diagonal Spearman correlation is $\rho = 0.79$, every benchmark pair is positive, and the economic benchmarks correlate with the academic core almost as strongly as with one another (Figure~\ref{fig:eda}a). A dominant general factor would produce this near-block structure, and the structure analysis must explain it, since a strong positive manifold is equally consistent with one underlying capability and with several capabilities that improve together over time. Scores rise steeply and jointly with release date (Figure~\ref{fig:eda}b), so release date bears on any claim that two benchmarks measure the same thing; the saturation literature finds saturation rising with benchmark age \citep{akhtar2026}, and we therefore adjust for date before reading the factor structure.

Most benchmarks are bounded accuracies in $[0, 1]$, and several are strongly right-skewed, with CritPt at Fisher skewness 3.8, HLE at 1.5, and \tauthree{} at 0.9, their scores piling up near the floor (Figure~\ref{fig:dist}). Two benchmarks are unbounded, GDPval on an Elo scale and AA-Omniscience on a penalised score, and both sit closer to symmetric. Bounded and skewed scales can manufacture or hide structure through the metric alone \citep{schaeffer2023}, which motivates the logit-transform check in Appendix~\ref{app:robust}; that check confirms that the first-factor share is no artefact of the raw scale. The low-scoring tails hold weak models whose scores are low, so we retain every row and handle the boundedness through the transform, trimming no outliers.

The dimensionality evidence behind H1, the scree plot and the parallel analysis, is Figure~\ref{fig:scree} in Appendix~\ref{app:derivations}; the first principal component accounts for 79.4\% of total score variance, and only the first observed eigenvalue exceeds its random-data 95th percentile.

\section{Hyperparameter grid and selections}
\label{app:hyper}
\emph{Supports Section~\ref{sec:task2} and the robustness claim of Section~\ref{sec:robust}} that the H4 verdict does not hinge on a single tuning choice. The inner grid searched ridge $\alpha \in \{0.03, 0.1, 0.3, 1, 3, 10, 30\}$, elastic-net $\alpha \in \{0.03, 0.1, 0.3, 1\}$ with $\ell_{1}$ ratio in $\{0.2, 0.5, 0.8\}$, random-forest depth in $\{4, \text{none}\}$ with 300 trees, and gradient-boosting depth in $\{2, 3\}$ with learning rate in $\{0.05, 0.1\}$. The ridge and elastic-net grids span three orders of magnitude of regularisation strength on a log scale, the standard practice for a regularisation path \citep{hastie2009}, and the tree grids bracket the shallow trees appropriate for $n = 96$ and $k + 4$ predictors. Table~\ref{tab:hyper} reports the modal selection per rung; the modal ridge $\alpha = 0.03$ sits at the weakly regularised end of the grid, and the H4 verdict is unchanged across the sweep of $k$ in Appendix~\ref{app:robust}. The regression learners and cross-validation use scikit-learn \citep{pedregosa2011}; the factor analysis uses the \texttt{factor\_analyzer} package, and SHAP its reference implementation \citep{lundberg2017}.

\begin{table}[!htb]
\caption{Modal hyperparameters selected by the inner five-fold grid search, reported as the plurality choice across all twelve leave-one-benchmark-out targets with the agreement count. The H4 verdict rests on fold-averaged out-of-fold errors and on no single setting.}
\label{tab:hyper}
\centering
\footnotesize
\setlength{\tabcolsep}{3pt}
\resizebox{\textwidth}{!}{%
\begin{tabular}{lllr}
\toprule
Rung & Learner & Modal hyperparameters & Agreement \\
\midrule
(i) timing / scale & random forest & \texttt{max\_depth=4}, \texttt{min\_samples\_leaf=3}, \texttt{n\_estimators=300} & 8/12 \\
(ii) mean index & ridge & \texttt{alpha=0.03} & 6/12 \\
(iii) first factor & elastic net & \texttt{alpha=0.03, l1\_ratio=0.2} & 5/12 \\
(iv) $k$ factors & ridge & \texttt{alpha=0.03} & 6/12 \\
(v) $k$ factors + covariates & ridge & \texttt{alpha=1} & 6/12 \\
\bottomrule
\end{tabular}}
\end{table}

\section{Deviations from the pre-specified design}
\label{app:deviations}
\emph{Supports the reproducibility claims throughout.} A study that fixes its analysis plan in advance owes the reader an explicit account of where the delivery departed from it. Six departures matter. First, the plan's primary unit was one row per base model at its best default configuration. The analysis instead uses all 421 configurations; the deduplicated base-model dataset appears in Appendix~\ref{app:robust} as planned check R1, and the configuration level serves as the primary. Second, the plan's factor-count rule is parallel analysis, and it retains one factor. The three-factor solution that exhibits economic separation is a pre-declared exploratory over-extraction, and we report H3 as not supported under the plan in consequence. Third, the taxonomy places Terminal-Bench Hard in scientific coding, while the plan grouped it with Terminal-Bench v2.1 as economic. The date-adjusted loadings show Hard at 0.82 on the economic factor, so the assignment is contestable, and we disclose the split openly. Fourth, the delivered robustness checks depart from the planned suite. We label them distinctly below, and we do not renumber a different battery onto the planned names. Fifth, the plan set the two inclusion floors as indicative values for confirmation at the coverage audit; the audit confirmed both unchanged, and no score was used to choose them. The plan also targeted 180 to 220 base models, where coverage across all twelve benchmarks delivered 96 complete cases and 89 deduplicated base models. Sixth, the plan specifies $B = 1{,}000$ bootstrap resamples, names both the timing baseline and the single index for H4, and fixes no learner for the contrast; we use $B = 2{,}000$, test against the single index as the stricter baseline, and report ridge with the other three learners in check C8.

Two further differences are presentational and are recorded for readers comparing the paper with the deposited plan. The plan organised the work under two questions, structure and prediction; the paper splits the structural question so that each hypothesis has its own research question, with no change to any hypothesis or test. And the H2(ii) fifteen-point threshold was calibrated to a reading of \citet{kearns2026} in which a scale correction lowered a dominant factor's share from 72\% to 41\%. That reading was mistaken, and Section~\ref{sec:related} gives the corrected account; the threshold itself was fixed in advance and has not been moved.

\section{Clustering of models and benchmarks}
\label{app:clustering}
\begin{figure}[!htb]
\centering
\includegraphics[width=\textwidth]{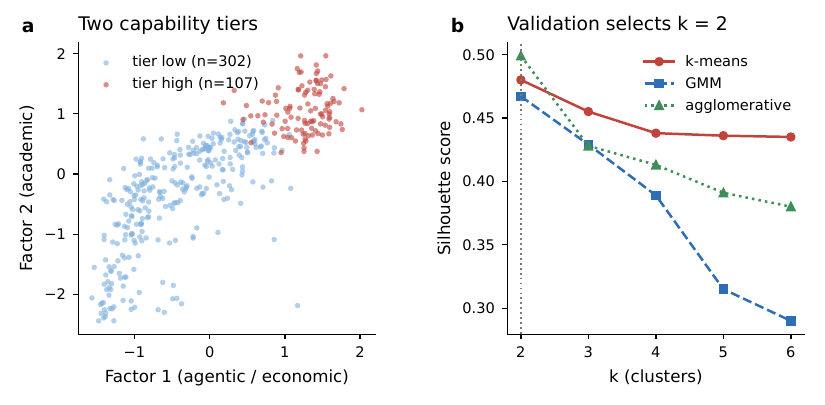}
\caption{\textbf{Model clustering on the dense grid ($n = 409$).} (a) Models in the space of the first two factors, coloured by the two-cluster agglomerative solution. The factor axes are the dense-grid solution fitted largely without the economic columns, so they track overall capability, and the two clusters mark tiers of it. (b) Silhouette scores by $k$; all three algorithms agree on $k = 2$.}
\label{fig:clusters}
\end{figure}

\emph{Supports Section~\ref{sec:task1}.} A factor model can impose structure the raw data do not support, so we sought a check that does not depend on the retention rule, and clustering in factor-score space corroborates the low dimensionality. We cluster models in factor-score space with $k$-means, Gaussian mixtures, and agglomerative Ward linkage, selecting the cluster count by silhouette \citep{rousseeuw1987} and the gap statistic \citep{tibshirani2001}; three algorithms guard against a clustering that reflects one algorithm's inductive bias. All three prefer $k = 2$ by silhouette, at 0.480 for $k$-means, 0.499 for agglomerative, and 0.467 for the Gaussian mixture (Figure~\ref{fig:clusters}b). The gap statistic increases monotonically in $k$ and finds no interior optimum, so it is inconclusive on data with a strong general factor and we set it aside. The two clusters mark capability tiers along one axis of overall strength. A larger, earlier, lower-scoring group of 302 models, with median release 2025-09 and mean Intelligence Index 13.7, separates from a smaller, newer, higher-scoring group of 107 models, with median release 2026-03 and mean index 37.1. The split sorts models by how much capability they hold, along the same date-driven axis (Figure~\ref{fig:clusters}a).

\begin{figure}[!htb]
\centering
\includegraphics[width=0.90\textwidth]{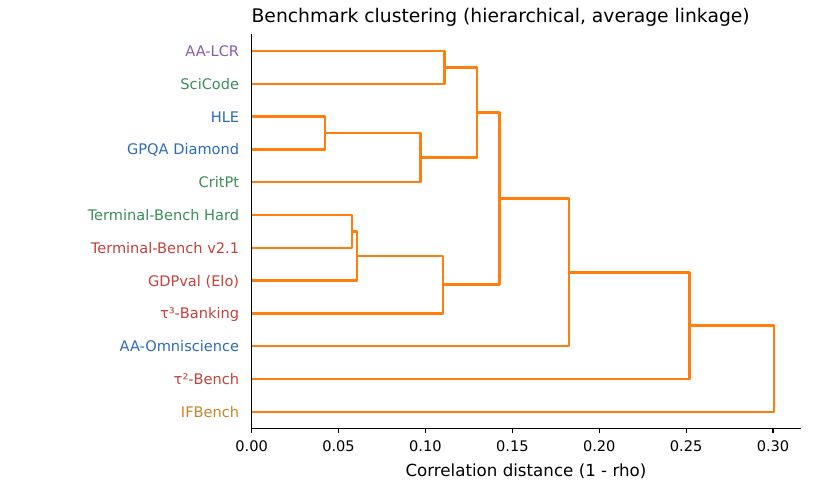}
\caption{\textbf{Benchmark clustering by correlation distance} ($1 - \rho$, average linkage). Ten of the twelve benchmarks, including three of four economic benchmarks, fall in one block, with \tautwo{} and IFBench branching separately.}
\label{fig:dendro}
\end{figure}

Clustering the benchmarks separately by correlation distance $1 - \rho$ under average linkage places ten of twelve, and three of four economic benchmarks, in one block, with \tautwo{} and IFBench branching separately (Figure~\ref{fig:dendro}).

\section{Robustness checks}
\label{app:robust}
\emph{Supports Section~\ref{sec:robust}.} We first report the one planned check that bears directly on the deduplication concern, then the additional checks actually delivered.

\paragraph{Planned check R1 (configuration versus deduplicated).} We collapse the 421 configurations to one row per base model, keeping the best default configuration as the plan specified. We operationalise it as the highest-intelligence-index row for each base model, after stripping reasoning, effort, and preview suffixes from the model name. Of the resulting base models, 89 carry the complete battery. On this deduplicated grid the first-factor share rises to 90.2\% and the date-adjustment drop rises to 24.1 points, and the four economic loadings remain above 0.40. Deduplication therefore strengthens H1 and H2(ii) and leaves the exploratory economic pattern intact, and the configuration-level analysis is the conservative choice for the structural headline. We also re-run the H4 predictive test on this deduplicated grid, where the resampling unit is a distinct base model. The pooled economic-block improvement of the $k$-factor representation over the single index is $+0.038$ with a 95\% interval of $[+0.020, +0.056]$. It matches the configuration-level $+0.037$ and again excludes zero, so the predictive gain does not depend on treating near-duplicate configurations as exchangeable.

\paragraph{Delivered checks C1--C8.} (C1) A rank-based factor analysis raises the first-factor share to 94.5\%, and (C2) a logit transform of the bounded accuracy benchmarks raises it to 91.8\%, so H1 is not a Pearson artefact. (C3) On the 58-model compute-known subsample, the date-only adjustment drops the first-factor share by 16.5 points, and adding log-compute jointly gives a smaller 9.3-point drop, so compute does not deepen the temporal correction beyond date on the same models. (C4) The three clustering algorithms agree on $k = 2$. (C5) The predictor-ladder ordering holds on both the economic block and all twelve targets. (C6) The dominant-factor date fit does not depend on the link function, since a logistic fit gives $\Rsq = 0.505$ against an ordinary-least-squares $\Rsq = 0.477$, and both clear 0.30 with the first factor ranked strongest, so H2(i) does not depend on that modelling choice. (C7) The $k$-factor predictive advantage does not depend on the exploratory factor count. Sweeping the rung-(iv) predictor over $k \in \{2, 3, 4, 5\}$, the improvement over the single index is positive at every count with a bootstrap interval excluding zero, reading $+0.021$ at $k = 2$, $+0.037$ at the $k = 3$ we use, $+0.058$ at the criterion's $k = 4$, and $+0.059$ at $k = 5$, with pooled test $\Rsq$ rising from 0.79 to 0.83 across the range. The gain increases with $k$, so the $k = 3$ we adopt is conservative for this test relative to the criterion's $k = 4$. (C8) The $k$-factor advantage depends on the learner family. Fixing each learner on both rungs, the pooled economic $\dmse$ is $+0.037$ for ridge and $+0.036$ for elastic net, positive on all four targets for both, and $-0.23$ for random forest and $-0.18$ for gradient boosting, negative on all four. The two tree learners fit the economic block worse on both rungs (test RMSE 0.49 and 0.53 on the index rung against ridge's 0.47) and degrade further with three factor inputs on 96 models, so the advantage holds for the linear learners that fit best and reverses for the tree learners; intervals were computed for ridge only.

The planned checks not run are R3 (FIML and MICE missingness sensitivity), R4 (replication of the single-factor foil then correction), R5 (reasoning-effort variants as a second scale axis), and R6 (the taxonomy swap test). Each was deferred for space; R5 and R6 bear on the effort and taxonomy departures above and come next.

\section{Explainability and error analysis}
\label{app:explain}
\begin{figure}[!htb]
\centering
\includegraphics[width=0.90\textwidth]{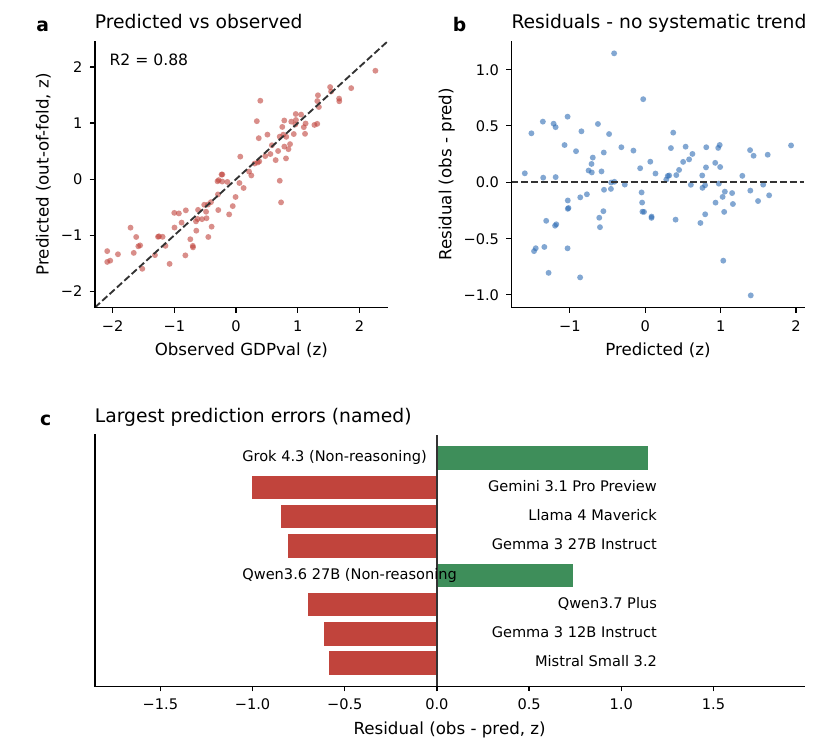}
\caption{\textbf{Regression diagnostics for the GDPval $k$-factor ridge model, out-of-fold.} (a) Predicted versus observed, on the identity line with $\Rsq = 0.88$. (b) Residuals versus predicted show no systematic trend. (c) The eight largest residuals, named. These are interpretable outliers, and none signals misspecification.}
\label{fig:regdiag}
\end{figure}

\emph{Supports Section~\ref{sec:h4}.} To interpret the predictive gain, we open the models fitted to the largest economic target. GDPval is predicted best by a $k$-factor ridge with out-of-fold $\Rsq$ 0.88 (Figure~\ref{fig:regdiag}). SHAP attribution on the gradient-boosting learner fitted at the same rung ranks the agentic factor F1 first, followed by the reasoning flag and the hard-reasoning factor F3, with scale and open-weights status contributing little (Figure~\ref{fig:lobo}c). The ordering is informative in the light of prior work, since \citet{ruan2024} find agentic performance forecastable from a low-dimensional capability space, and \citet{ilic2024} find parameter count correlated with the general factor at about 0.6; here, once the factors are in the model, parameter count and open-weights status add almost nothing, so the scale signal that prior work reports reaches GDPval through the factors alone. The predicted-versus-observed plot lies on the identity line, and the residuals show no systematic trend (Figure~\ref{fig:regdiag}a,b). The largest residuals are interpretable outliers. They include two non-reasoning variants that outperform their economic predictions, Grok 4.3 Non-reasoning at $+1.14$ and Qwen3.6 27B Non-reasoning, and six models that under-perform theirs, led by Gemini 3.1 Pro Preview and Llama 4 Maverick (Figure~\ref{fig:regdiag}c). The pattern fits GDPval rewarding agentic behaviour that the other benchmarks capture only indirectly.

\section{Data provenance and reproducibility}
\label{app:provenance}
\emph{Supports Section~\ref{sec:data} and the reproducibility claims throughout.} The primary data set is the Artificial Analysis model leaderboard \citep{aa2026}, captured 2026-07-06 as a raw HTML snapshot and pinned by the SHA-256 hash recorded in the shipped manifest (\texttt{aa\_snapshot\_manifest.json}), \texttt{6f19f8f08befaaf14cb2d995}\allowbreak\texttt{2e379404c39f246b52d7ae4d1c2aa44795404de6}. The Data API refused an unauthenticated request, so we captured the public page and stored the exact bytes. Every downstream number is therefore reproducible regardless of live-page drift \citep{biderman2024}. The secondary data set is Epoch AI Notable AI Models \citep{epoch2026}, released under CC-BY 4.0 and used only for the scale robustness check. The analysis plan was fixed before the data were captured and before any analysis was run, and is deposited at \url{https://doi.org/10.17605/OSF.IO/VD34J}; the deposit is retrospective, and the note there records the timeline, the plan's checksum, and the limits of that evidence. The recorded timeline is that the plan was first saved at 14:18 UTC on 6 July 2026, the leaderboard snapshot was captured at 15:17 and the first analysis artefact was produced at 15:23, so the plan precedes the data by 59 minutes and the analysis by 65; the plan's three recorded copies are byte-identical, and the timestamps come from the working environment's own file record, which is weaker evidence than a registry timestamp would be. Both snapshots, every result table, and the analysis notebook are released at \url{https://github.com/louisyzhu/frontier-ai-economic-validity}. Reproducibility is tiered, and each result belongs to one tier. The structural results, the SHAP attribution, the deduplication checks, and the prediction ladder are \emph{recomputed} from the pinned inputs; twelve result tables and four appendix figures are \emph{regenerated} from archived outputs, which the notebook reads and prints without recomputing. The pipeline is a single notebook that runs top to bottom from the pinned snapshots with no manual step, no live network call and no path outside the repository, and all four learners run by default. It regenerates the figures that carry the results, Figures~\ref{fig:loadings}, \ref{fig:lobo}, \ref{fig:eda}, \ref{fig:clusters} and~\ref{fig:regdiag}, while the remaining figures and twelve result tables listed in the repository are shipped without being recomputed within it; the notebook prints every value it reads, so each remains checkable against its own output. The SHAP attribution behind Figure~\ref{fig:lobo}c and the R1 deduplication with its H4 re-run verify themselves against the tables cited here and reproduce exactly.

The prediction ladder is refitted on every run and written alongside the canonical table, so no value reported here can be altered by re-execution. The notebook then recomputes the quantities this paper prints, the per-target and pooled economic $\dmse$ and the five economic-block rows of Table~\ref{tab:ladder}, and compares them with the canonical values at a tolerance of $5 \times 10^{-4}$, half a unit at the third decimal and so at the precision reported here; it separately verifies that the best-learner ranking behind the table's learner column is unchanged. Those quantities reproduce in every printed digit on both machines we have tested. Individual ladder cells are reported without being enforced, because one rung is not platform-invariant. The maximum-likelihood solution has a weakly determined leading direction for this correlation structure, so the single-factor rung can settle on a slightly different point under a different linear-algebra backend, and gradient-boosted trees carry that difference most visibly because a split threshold either flips or does not. The largest deviation we observed was $7.2 \times 10^{-4}$, reaching the $k$-factor rung only through tree learners on a non-economic target, while the mean-index and $k$-factor ridge fits from which every reported effect and interval is built agreed to about $10^{-7}$. The divergence is a property of the platform and of that rung and leaves the results untouched. The library benchprobe v0.1.0 \citep{benchprobe2026}, extracted from this notebook and archived at \url{https://doi.org/10.5281/zenodo.22705351}, recomputes the structure results and the ladder point estimates from the same pinned snapshot by one command, statistically reproduces the bootstrap interval and the $k$-sweep under a fresh random stream, and regenerates the clustering table and three further fields from the archive; its ledger records the archived value, the recomputed value and the tolerance for every quantity, and 23 of its 44 acceptance tests lock this paper's numbers. Two of its computations reconstruct steps the archive does not ship, the four-parameter logistic behind the date $\Rsq$ and the mean-of-RMSE pooling behind the ladder table, and both are recorded as reconstructions in its decision record.

\end{document}